\documentclass[runningheads]{llncs}
\usepackage{graphicx}

\usepackage{tikz}
\usepackage{comment}
\usepackage{amsmath,amssymb} 
\usepackage{color}
\usepackage{algorithm}
\usepackage{algorithmic}
\usepackage{amsmath}
\usepackage{multirow} 
\usepackage{booktabs} 
\usepackage{array} 
\usepackage{amssymb} 
\usepackage{pifont} 
\newcommand{\cmark}{\ding{51}}%
\newcommand{\xmark}{\ding{55}}%
\newcommand{\ul}{\underline}
\usepackage{tabularx}
\usepackage{color}
\usepackage[misc,geometry]{ifsym}
\usepackage[colorlinks, linkcolor=red, anchorcolor=blue, citecolor=green]{hyperref}

\usepackage[accsupp]{axessibility}  

\begin{document}
\pagestyle{headings}
\mainmatter
\def\ECCVSubNumber{1712}  

\title{One-Shot Medical Landmark Localization by Edge-Guided Transform and Noisy Landmark Refinement}

\titlerunning{One-Shot Landmark Localization by Edge-Guided Transform \& Refinement}
%
\author{
Zihao Yin\inst{1} \and 
Ping Gong\inst{2} \and 
Chunyu Wang\inst{3} \and 
Yizhou Yu\inst{4} \and 
Yizhou Wang\inst{5,6}$^{\left(\textrm{\Letter}\right)}$ 
}
\authorrunning{Z Yin et al.}
%
\institute{
Center for Data Science, Peking University, Beijing, China \and
Deepwise AI Lab, Beijing, China \and
Microsoft Research Asia, Beijing, China \and 
The University of Hong Kong, Hong Kong \and
Center on Frontiers of Computing Studies, School of Computer Science, Peking University, Beijing, China \and
Inst. for Artificial Intelligence, Peking University, Beijing, China \\
\email{\{silvermouse, yizhou.wang\}@pku.edu.cn, gongping@deepwise.com, chnuwa@microsoft.com, yizhouy@acm.org}
}
\maketitle

\begin{abstract}
As an important upstream task for many medical applications, supervised landmark localization still requires non-negligible annotation costs to achieve desirable performance. Besides, due to cumbersome collection procedures, the limited size of medical landmark datasets impacts the effectiveness of large-scale self-supervised pre-training methods. To address these challenges, we propose a two-stage framework for one-shot medical landmark localization, which first infers landmarks by unsupervised registration from the labeled exemplar to unlabeled targets, and then utilizes these noisy pseudo labels to train robust detectors. To handle the significant structure variations, we learn an end-to-end cascade of global alignment and local deformations, under the guidance of novel loss functions which incorporate edge information. In stage II, we explore self-consistency for selecting reliable pseudo labels and cross-consistency for semi-supervised learning. Our method achieves state-of-the-art performances on public datasets of different body parts, which demonstrates its general applicability.
\keywords{Medical Landmark Localization, One-Shot Learning}
\end{abstract}

\section{Introduction}
Landmark localization is an essential step for many medical image applications, such as dental radiography~\cite{wang2016benchmark,chen2019cephalometric}, bone age assessment~\cite{escobar2019hand,gong2020towards}, vertebra labeling~\cite{payer2019integrating} and per-operative measurements~\cite{liu2020landmarks}. Although fully supervised methods~\cite{wang2020deep,li2020structured,liu2021polarized} achieve the state-of-the-art results, the required manual annotations take considerable cost and time. In contrast, given few exemplars and proper instructions, experts are ready to generalize these landmark concepts to unseen targets and annotate them accurately. This motivates us to explore the challenging task of one-shot medical landmark localization.

Besides the scarce supervision, significant differences in spatial structures between images for landmark datasets also increase the difficulty of this task. While they can be quite different in scale, orientation, or intensity due to patient positioning or imaging quality as in hand radiography~\cite{escobar2019hand}, there are also substantial variations in local structures, such as the front teeth in dental radiography~\cite{wang2016benchmark}. Furthermore, because of cumbersome acquisition procedures, medical landmark datasets are too expensive to collect in large numbers. Thus, the amount of unlabeled data available is often limited.

Considering these challenges, landmark localization in the low data regime is in urgent need and explored by~\cite{browatzki20203fabrec,yao2021one,zhou2021scalable}. 3FabRec~\cite{browatzki20203fabrec} is a method for few-shot face alignment. They first train an autoencoder for face reconstruction and then retask the decoder to heatmap prediction through fine-tuning on labeled sets. However, when having access to only one exemplar and hundreds of data as in our work, qualities of reconstructed images are too poor to perform landmark localization. Motivated by the recent success of contrastive learning, CC2D~\cite{yao2021one} proposes to detect target landmarks by first solving a self-supervised patch matching task and uses these pseudo labels to retrain new detectors. However, their method overlooks the global spatial relationships of landmarks and is prone to yield inaccurate predictions once overfitting to the local appearance of specific instances.  Conversely, DAG~\cite{li2020structured} employs graph convolution network (GCN) to capture topological constraints of landmarks. Few-shot DAG~\cite{zhou2021scalable} extends DAG to the few-shot (e.g., five-shot) setting. They report impressive results on several datasets but fail to converge under the extreme one-shot setting. 

\begin{figure*}[t]\centering\label{scheme}
\includegraphics[width=0.975\textwidth]{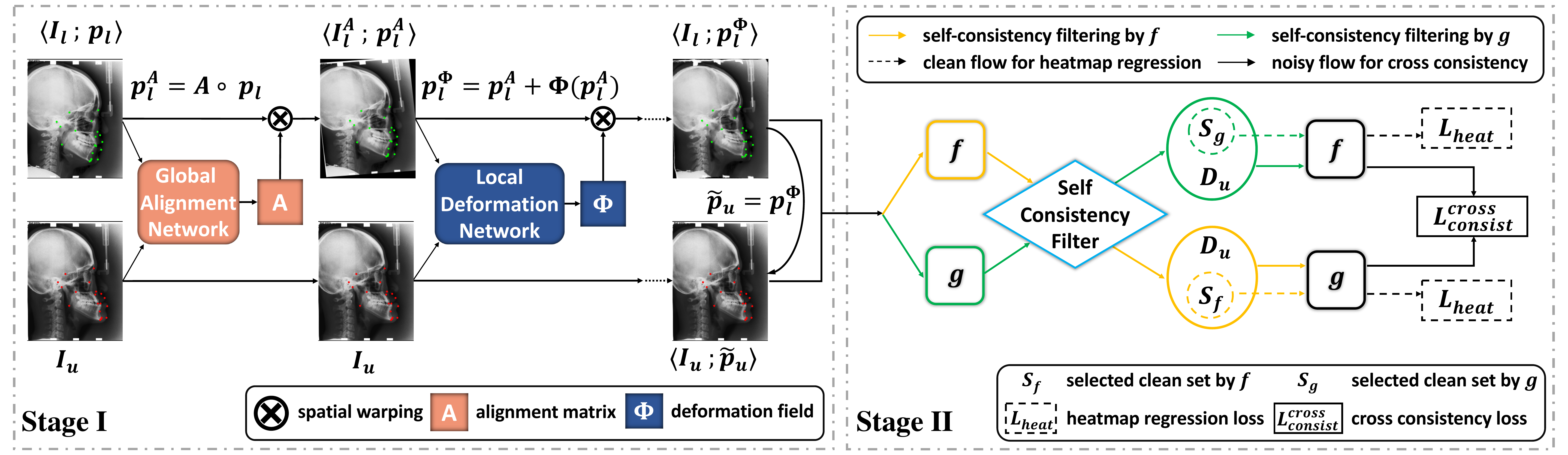} 
\caption{Overview of the proposed framework. In stage \uppercase\expandafter{\romannumeral1}, unsupervised registration is learned through an end-to-end cascade of global alignment and subsequent local deformations, which aims to predict pseudo landmarks $\tilde{p}_u$ by registering the labeled exemplar $I_l$ to unlabeled targets $I_u$. Inferred noisy landmarks are further refined in stage  \uppercase\expandafter{\romannumeral2}, where two robust landmark detectors $f,g$ are co-trained, by utilizing both self-consistency for sample selection and cross-consistency for semi-supervised learning}
\end{figure*}

In order to fully exploit the exemplar and available data, we propose to propagate the landmarks from exemplar to unlabeled images through registration, which not only considers global anatomical constraints, but also performs precise matching of local structures. In fact, the dense correspondence between instances learned by registration can be directly leveraged by landmark localization. Besides, registration can be learned efficiently without the need of a large amount of data to produce reasonable results. As shown in Fig.~\ref{scheme}, our novel two-stage framework first learns unsupervised registration from the labeled exemplar to unlabeled targets for inferring pseudo landmarks, and then
trains robust landmark detectors by exploiting consistency between clean annotations and noisy pseudo labels. For better adaptation to landmark localization, we make several non-trivial contributions to solve the following challenges.

First, there might be a chicken-and-egg issue if you want to infer landmarks for unseen targets through registration, since registration itself usually requires detected landmarks, either for alignment in pre-processing steps~\cite{balakrishnan2018unsupervised}, or to guide the learning process as extra structural information~\cite{lee2019image}. To avoid this dilemma, we decompose the total spatial transform into an end-to-end cascade of global alignment and local deformations. To facilitate the registration learning, powerful attention blocks~\cite{vaswani2017attention} are employed, including self-attention for capturing long-range dependencies and cross-attention for fusing multi-resolution features. 

Second, classical reconstruction terms for registration, which mainly consider image similarity based on the distribution of pixel values, are insufficient to constrain the structural consistency of landmarks. Hence we propose novel loss functions that incorporate edge information and landmark locations. Specifically, our reconstruction term further involves the masked similarities of edge structures around interested landmarks. Besides, since different anatomical parts tend to have different displacements, it is beneficial to relax the smoothness constraints of deformation field around boundaries.

Last, under the interference of certain nuisances (e.g., background, abnormal appearance), deformation learned by the low-level registration task can not perfectly capture the high-level semantic correspondence of landmarks. Thus, an essential second stage is introduced to refine noisy predictions with large biases. We train robust landmark detectors, utilizing self-consistency between different views of the same model for sample selection and cross-consistency between different views of different models for semi-supervised learning.

We conduct experiments on public medical landmark datasets of different body parts, including head, hand and chest. Our method consistently outperforms other baselines with notable margins and further narrows the gap with the supervised upper bound. To summarize, our contributions are three-fold:
\begin{enumerate}
\item We propose an unsupervised training strategy for inferring pseudo landmarks through registration, which learns an end-to-end cascade of global alignment and local deformations, with the guidance of novel loss functions incorporating edge information.
\item We introduce an effective scheme for training robust landmark detectors with noisy labels, which utilizes self-consistency for selecting reliable pseudo labels and cross-consistency for semi-supervised learning. 
\item We conduct experiments on public medical landmark datasets of different body parts. Results show our method stably advances the state-of-the-art for all three applications.
\end{enumerate}

\section{Related Work}
We briefly review most related works, including one-shot, few-shot and semi-supervised methods for landmark localization, and medical image registration. 
\subsection{One-shot and few-shot landmark localization} 
As mentioned, CC2D~\cite{yao2021one} is motivated by contrastive learning: features for the original patch and its randomly augmented counterparts at the same location are matched using cosine similarity. With the learned matching network and template patches, pseudo-labels are inferred for retraining a multi-task UNet~\cite{yao2020miss} from scratch. Despite their promising results on the cephalometric dataset~\cite{wang2016benchmark}, CC2D overlooks valuable global structure constraints~\cite{li2020structured}, making it difficult to handle multiple similar local structures, such as fingertips. Another interesting work, 3FabRec~\cite{browatzki20203fabrec} achieves impressive performance for few-shot face alignment. They first train an adversarial autoencoder for unsupervised face reconstruction, then fine-tune with interleaved layers to the landmark detection task with few labels. 3FabRec demonstrates great benefits of dense pixel-level pre-training for landmark localization. However, for medical applications where the amount of data is orders of magnitude less, it is rather difficult to achieve satisfactory results through pre-training of image reconstruction.
\subsection{Semi-supervised landmark localization} 
Recent advances can be divided into two streams: consistency-based approaches
\cite{honari2018improving,moskvyak2021semi} and synthetic image based approaches~\cite{qian2019aggregation,kumar2020s2ld}. The equivariant landmark transformation (ELT) constraint~\cite{honari2018improving} is built on the intuition that, given a transformed image, the model should produce similarly transformed landmarks. Semantic Consistency~\cite{moskvyak2021semi} encourages learning similar features for landmarks with the same semantics across images. Synthetic image approaches focus on generating desired training data. StyleAlign~\cite{qian2019aggregation} first disentangles face images to style and structure space, then transfers randomly sampled styles to images with known structures, greatly enriching training space.
\subsection{Learning-based deformable image registration} 
Our work is closely related to learning-based image registration~\cite{uzunova2017training,balakrishnan2018unsupervised,li2018non,lee2019image}. \cite{lee2019image} proposes to utilize detected landmarks as extra structural information to guide the training of registration. VoxelMorph~\cite{balakrishnan2018unsupervised} proposes to use a convolutional neural network (CNN) $g$ to learn the registration field $\Phi$ unsupervisedly, by optimizing the image similarity between the fixed and moving images, and the smoothness of $\Phi$. \cite{balakrishnan2018unsupervised} requires image pairs to be affinely aligned in the pre-processing step and then focus on the nonlinear correspondence. \cite{balakrishnan2018unsupervised} can further be extended as in DataAug~\cite{zhao2019data} for one-shot medical segmentation, by learning independent spatial and appearance transform for data augmentation.

\begin{figure*}[th]\centering
\includegraphics[width=.95\textwidth]{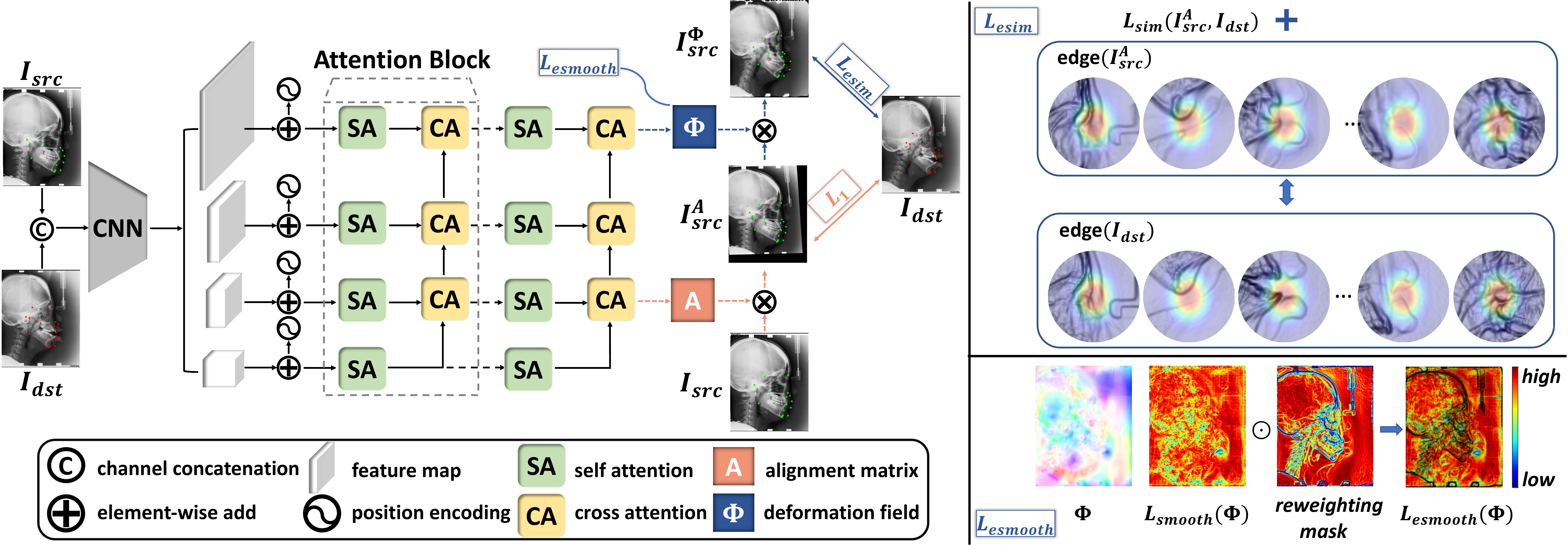}
\caption{The architecture of registration model consists of a CNN backbone and cascades of attention blocks. Red lines denote the flow for global alignment, while blue lines denote the flow for local deformation. Besides the image similarity, $\mathcal{L}_{esim}$ is introduced to further penalize local similarities of edge structures around inferred pseudo landmarks, masked by their gaussian heatmaps. $\mathcal{L}_{esmooth}$ relaxes the smoothness constraint around boundaries, to avoid mutual interference between different regions (e.g., anatomical parts, background \& foreground)}
\label{arch}
\end{figure*}

\section{Method}
Let $I_l, I_u$ be two images defined over a 2-D spatial domain $\Omega$. $I_l \in \mathbb{R}^{H\times W}$ is the labeled exemplar with $N$ annotated landmarks $p_l =\{(x_l^j, y_l^j)| j=1,\cdots, N\}$, and $I_u$ is the unlabeled target. $I_l$ and $I_u$ share similar appearance distribution, and landmarks $p_l$ are defined at locations with particular anatomical structures. Our goal is to learn a spatial transformation from $I_l$ to $I_u$ through registration, so that we can infer reliable landmarks for $I_u$ based on $p_l$, according to the equivariant property of landmarks.

\subsection{Reformulation of Classical Registration Framework}
First, in order to adapt to the downstream landmark localization, we reformulate the classical framework for unsupervised registration as follows:
\begin{align}\centering\label{vx_new}
    & \hat{A},\hat{\Phi} = \arg\min_{A, \phi}\mathcal{L}_{total}(I_{src}, I_{dst}\ |\  \tilde{p}_{src}), A = g_\theta(I_{src}, I_{dst}), \Phi = g_\theta(I_{src}^A, I_{dst})\\
    & \mathcal{L}_{total} = \mathcal{L}_{global}(I_{src}^A, I_{dst}) + \mathcal{L}_{local}(I_{src}^{\Phi}, I_{dst}\ | \  \tilde{p}_{src}^{\Phi}) + \mathcal{L}_{esmooth}(\Phi \ |\ I_{src}^{\Phi}) \\
    & I_{src}^{A} = A \otimes I_{src},\ I_{src}^{\Phi} = \Phi \otimes I_{src}^{A},\ \tilde{p}_{src}^A = A\circ \tilde{p}_{src},\ \tilde{p}_{src}^{\Phi} = \tilde{p}_{src}^A + \Phi(\tilde{p}_{src}^A) \label{pts_trans}
\end{align}
where we learn an end-to-end cascade of global affine alignment $A\in \mathbb{R}^{2\times3}$ and local deformation $\Phi\in \mathbb{R}^{H\times W\times2}$. During training, $\left\langle I_{src}, I_{dst} \right\rangle$ is a pair of images randomly sampled from the mini-batch. For inference, $I_{src}$ is fixed as the exemplar $I_l$ and we set $I_{dst}$ as the unlabeled target $I_u$ to predict pseudo landmarks $\tilde{p}_u$. The local similarity $\mathcal{L}_{local}$ is conditioned on $\tilde{p}_{src}^{\Phi}$ (Eq.~\ref{pts_trans}), where $\tilde{p}_{src}$ is an exponential moving average of $e$-th epoch prediction $\tilde{p}_{src}^e$ after certain epoch $M$:
\begin{equation}\label{ema}
\tilde{p}_{src} = \tau \tilde{p}_{src} + (1 - \tau)\tilde{p}_{src}^e,\ e >= M 
\end{equation}
The smoothness constraint $\mathcal{L}_{esmooth}$ is extraly conditioned on $I_{src}^{\Phi}$, to utilize its edge information, which will be discussed in Eq.~\ref{esmooth}. 

\subsection{Edge-Guided Global \& Local Transform}
As in Fig.~\ref{arch}, our registration model $g_\theta$ consists of a CNN backbone for feature extraction and several attention blocks for feature fusion.

We adopt the HRNet18~\cite{sun2019deep} as backbone, which takes a channel-wise concatenation of $I_{src}$ and $I_{dst}$ as input, and outputs four resolutions of feature maps $F_i \in \mathbb{R}^{C\times {H_i} \times {W_i}}$ with stride $S_i$ ($H_i = \frac{H}{S_i}, W_i = \frac{W}{S_i}, S_i \in \{32,16,8,4\}, i\in\{1,2,3,4\}$). For each $F_i$, we use a $1\times1$ convolution to change its channel dimension into a unified value $d_{model}$  and then reshape it to $\mathbb{R}^{d_{model} \times H_iW_i}$, since the attention blocks expect a sequence as input. Following \cite{parmar2018image,carion2020end}, we adopt a concatenation of two 1D learned positional encodings as the recovered order information. 

Because of the significant structural differences between $I_{src}$ and $I_{dst}$, distance between their corresponding pixels could be large. To capture these long-range dependencies, we furnish our model with powerful transformer layers~\cite{vaswani2017attention}:
\begin{align}
    & \textrm{Attention}(Q, K, V) = \textrm{softmax}(\frac{QK^T}{\sqrt{d_k}})V \\
    & \textrm{MultiHead}(Q, K, V) = \textrm{Concat}(\textrm{H}_1, \cdots, \textrm{H}_h)W^O \\
    & \textrm{H}_i = \textrm{Attention}(Q W_i^Q, KW_i^K, VW_i^V) 
\end{align}
where $Q, K, V$ are embeddings of the same feature map, projected into different spaces for the self-attention module (SA). It is also crucial to fully exploit features from multiple resolutions for registration learning. Thus, we further utilize the cross-attention module (CA) for cross-resolution feature fusion, where $K, V$ are embeddings of the feature map from branch of lower-resolution than $Q$. For each layer of attention blocks, we stack both one SA module and one CA module following feature maps from each resolution, which progressively incorporates the information from lower resolutions to higher resolutions, and enables the model to gradually figure out the optimal transformation from $I_{src}$ to $I_{dst}$. 

\subsubsection{Global Transform} Since high-level features are sufficient to capture global relationships, features of lower resolutions (i.e., $F_1, F_2$) are passed into transformer layers and we use the aggregated representation of the last output state for affine estimation. It is then passed to a two-layer MLP and tanh function to obtain $o\in \mathbb{R}^6$, each element of which denotes the relative changes in translation $(t_x, t_y)$, scale $(s_x, s_y)$, rotation $\alpha$, and shear $\beta$:
\begin{align}
    o =& \tanh(\textrm{MLP}(H)) \\
    t_x =& o_1,\ s_x = 1 + o_3 * sf_x,\ \alpha = o_5 * rot \\
    t_y =& o_2,\ s_y = 1 + o_4 * sf_y,\ \beta = o_6 * sh
\end{align}
where $sf_x, sf_y, rot, sh$ are four hyper parameters controlling transformation intensities. Then the affine matrix $A\in \mathbb{R}^{2\times 3}$ can be computed as follows: 
\begin{center}
    $\begin{pmatrix}
    s_x\cos\alpha  & s_x(\cos\alpha\tan\beta + \sin\alpha)  & t_x \\
    -s_y\sin\alpha & s_y(-\sin\alpha\tan\beta + \cos\alpha) & t_y 
    \end{pmatrix}$
\end{center}
which is a composite of several basic transforms. Then $I_{src}^{A}$ is obtained through differentiable bilinear interpolation based on the spatial transformer module~\cite{jaderberg2015spatial}. 

\subsubsection{Local Deformation Field}
Based on the global transform $A$, features are again extracted for $\left\langle I_{src}^A, I_{dst}\right\rangle$. For a more precise pixel-level correspondence, we mainly utilize high-resolution features to predict a deformation field $\Phi \in \mathbb{R}^{H\times W\times2}$. Since global transform already roughly aligns two images, the target pixel for a certain location of $I_{src}^A$ is more likely to be within a local neighborhood in $I_{dst}$. Thus, for high-resolution features (i.e., $F_3, F_4$), we can reduce their spatial scale through reshaping them from $C\times H_i W_i$ to $R^2$ sequences (e.g., $R=4$) with size $C\times \frac{H_i W_i}{R^2}$ and perform attention mechanism on each sequence. 

The final hidden state from the highest-resolution branch is fed into a linear layer for regressing displacements $(\Delta{x}, \Delta{y})$, which implies that $I_{src}^G(x$
$+\Delta{x}, y + \Delta{y})$ corresponds to $I_{dst}(x, y) \ (x\in\{1,\cdots,h\}; y\in\{1,\cdots,w\})$ . With the deformation field $\Phi$, we further improve $I_{src}^A$ to $I_{src}^\Phi$ through bilinear interpolation. 

\subsubsection{Loss \& Training Strategy}
For learning of the challenging unsupervised deformation, we apply the local deformation cascade iteratively for $N_\Phi$ times. Thus, the total loss is a weighted combination of one global similarity $\mathcal{L}_{global}$, local similarity $\mathcal{L}_{local}$ and regularization $\mathcal{L}_{reg}$ of $\Phi_i$ for each step.
\begin{align}\centering
& \mathcal{L}_{stage_{\uppercase\expandafter{\romannumeral1}}} = \mathcal{L}_{global}(I_{src}^A, I_{dst}) + \lambda_1\sum_{i=1}^{N_\Phi}{[\mathcal{L}_{local}(I_{src}^{\Phi_i}, I_{dst}) + \mathcal{L}_{reg}(\Phi_i)]} \label{loss_st1}\\
& \mathcal{L}_{reg}(\Phi_i) = \mathcal{L}_{smooth}(\Phi_i) + \lambda_2\mathcal{L}_{inv}(\Phi_i) + \lambda_3\mathcal{L}_{syn} \label{loss_reg}
\end{align}
Our ultimate goal is to find the discrete semantic correspondence between landmarks, instead of image registration. Similarity terms during the deformation process should gradually weaken impacts of appearance and focus more on structural information. We simply use $\mathcal{L}_1$ for $\mathcal{L}_{global}$. While for $\mathcal{L}_{local}$, we propose to enhance the original image similarity with an additional masked similarity term between their edge maps as follows:
\begin{align}\centering\label{esim}
& \mathcal{L}_{esim} = \mathcal{L}_{sim}(I_{src}^\Phi, I_{dst}) + \mathcal{L}_{sim}(\textrm{edge}(I_{src}^\Phi), \textrm{edge}(I_{dst})) \odot \textrm{Mask}(p_{src}^\Phi) \\
& \textrm{Mask}(p) = \frac{1}{N}\sum_{i=1}^{N}{\exp(-\frac{1}{2\sigma^2}||u - p_i||^2)},\ p\in \mathbb{R}^{N\times 2},\ \forall u \in \Omega
\end{align}
where \textrm{edge$(\cdot)$} denotes the operator for edge detection (e.g., sobel) and \textrm{Mask$(\cdot)$} generates the averaged gaussian heatmaps with fixed deviation $\sigma$, centered on landmarks $p_{src}^\Phi$. $\mathcal{L}_{esim}$ considers the similarity between not only image pairs, but also their local edge maps around landmarks, which is beneficial to enforce the structural consistency of landmarks across images. We adopt the robust structural similarity (SSIM)~\cite{wang2003multiscale} as $\mathcal{L}_{sim}$ in Eq.~\ref{esim}.

Besides, we observe that different anatomical parts tend to exhibit different displacements.
Thus, the original smoothness constraint $\mathcal{L}_{smooth}$ (Eq.~\ref{smooth}), which penalizes the approximated spatial gradients of $\Phi$ for all pixels, should be relaxed around the boundaries of these parts. We propose to re-weight $\mathcal{L}_{smooth}$ based on the magnitudes of detected edge vectors as in Eq.~\ref{esmooth}:
\begin{align}\centering
& \mathcal{L}_{smooth}(\Phi) = ||\nabla\Phi(u)||^2,\ \forall u\in\Omega \label{smooth} \\
& \mathcal{L}_{esmooth}(\Phi | I_{src}^\Phi) = \mathcal{L}_{smooth}(\Phi) \odot \exp(-||\textrm{edge}(I_{src}^\Phi)||^2 / T) \label{esmooth}
\end{align}
where $T$ is a hyper parameter to control the sharpness of distribution. In this way, weights for boundaries is lowered by their large magnitudes in edge maps and attenuates the mutual interference between different parts. As in~\cite{balakrishnan2018unsupervised}, we also adopt $\mathcal{L}_{inv}$ to enforce the invertibility of $\Phi$. For ease of early optimization, we introduce $\mathcal{L}_{syn}$, where we apply random perspective transform to $I_{src}$ and use synthetic pairs with known correspondence to supervise the learning of $\Phi$.

\subsection{Stage \uppercase\expandafter{\romannumeral2}: Noisy Landmark Refinement}
With the learned spatial correspondence from the exemplar $(I_l, p_l)$ to $I_u$, we can infer pseudo landmarks $\tilde{p}_u$ for $I_u$. $h(\cdot)$ computes gaussian heatmaps with fixed standard deviation $\sigma$ centered on landmark locations. Instead of simply train a new landmark detector with these pseudo-labels, as~\cite{yao2021one} did, we propose a robust learning scheme for noisy landmarks, coined as \textit{Consistency Co-Teaching} (C2T).

As shown in Alg.~\ref{alg:coteach}, C2T builds on the seminal co-teaching framework~\cite{han2018co}. Two
networks $f$ and $g$ are co-trained to select relatively clean samples $S_f$ and $S_g$ (line 5-6) for the other network (line 7-9), leveraging the small-loss assumption~\cite{Zhang2017UnderstandingDL,arpit2017closer,han2018co}. C2T's main novelties are two ingredients: self-consistency filtering $\mathcal{L}_{filter}$ (Eq.~\ref{lheat}-\ref{lfilter}) and cross-consistency loss $\mathcal{L}_{con}^{cross}$ (Eq.~\ref{lcon}), which not only stabilize training, but also boost performance.

\begin{align}\centering
    &\mathcal{L}_{heat}(f, x, h(\tilde{p}_x)) = ||f(x) - h(\tilde{p}_x)||_2\label{lheat}\\
    &\mathcal{L}_{con}^{self}(f, x) = ||f(T_h(x)) -  T_{e\rightarrow h}(f(T_e(x)))||_2\label{lselfcon} \\
    &\mathcal{L}_{filter}(f, x, h) = \mathcal{L}_{heat}(f, x, h) + w\mathcal{L}_{con}^{self}(f, x)\label{lfilter} 
\end{align}

As pointed out in \cite{yu2019does}, in later training epochs, two networks trained by co-teaching could harmfully converge to a consensus. To keep $f$ and $g$ healthily diverged, we introduce $\mathcal{L}_{con}^{self}$(Eq.~\ref{lselfcon}) into $\mathcal{L}_{filter}$. $(T_e, T_h)$ is a pair of easy (weak augmentations) and hard (strong augmentations) views of the original images. Intuitively, if heatmaps predicted by $f$ ($g$) for $x$ are equivariant to different transformations~\cite{honari2018improving}, its pseudo label is more likely to be clean.

\begin{algorithm}[tb]
\caption{\bf{C2T: Consistency Co-Teaching}}
\label{alg:coteach}
\textbf{Input}: $L:\{(I_l, H_l)\}, U:\{(I_u, \hat{H}_u)\}_{u=1}^{N_u}$\\
\textbf{Parameter}: $f, g$: landmark detectors; $\epsilon$: filter rate\\
\textbf{Output}: $f,g$: landmark detectors
\begin{algorithmic}[1] 
\STATE \textbf{Shuffle} $L$ and $U$.
\FOR{$T=1, \cdots, T_{max}$}
\STATE \textbf{Fetch} mini-batch $D_l, D_u$ from $L, U$ respectively.
\STATE \textbf{Sample} random augmentations $T_e, T_h$ and compute transform $T_{e\rightarrow h}$.
\STATE $S_g = \mathop{\arg\min}\limits_{D'\ge \epsilon|D_u|}{\sum\limits_{(x,h)\in D'}L_{filter}(g, x, h)}, D'\subset D_u$.
\STATE $S_f = \mathop{\arg\min}\limits_{D'\ge \epsilon|D_u|}{\sum\limits_{(x,h)\in D'}L_{filter}(f, x, h)}, D'\subset D_u$.
\STATE $\mathcal{L}_f^{update} = \sum_{(x, h)\in D_l \cup S_g}\mathcal{L}_{heat}(f, x, h) + \sum_{(x,h)\in D_l \cup D_u}\mathcal{L}_{con}^{cross}(f, g, x) $.
\STATE $\mathcal{L}_g^{update} = \sum_{(x, h)\in D_l \cup S_f}\mathcal{L}_{heat}(g, x, h) + \sum_{(x,h)\in D_l \cup D_u)}\mathcal{L}_{con}^{cross}(g, f, x) $.
\STATE \textbf{Update} $f$ by $\nabla_{f}\mathcal{L}_f^{update}$ and $g$ by $\nabla_{g}\mathcal{L}_g^{update}$.
\ENDFOR 
\STATE \textbf{Return} $f,g$
\end{algorithmic}
\end{algorithm}
Instead of simply discarding filtered samples as in~\cite{han2018co}, we involve them along with selected samples, in $\mathcal{L}_{con}^{cross}$ (Eq.~\ref{lcon}) to explore the consistency between
\begin{align}\label{lcon}
    \mathcal{L}_{con}^{cross}(f, g, x) = |&|f(T_h(x)) - T_{e\rightarrow h}(g(T_e(x)))||_2 
\end{align}
predictions of different models on different views. Motivated by semi-supervised methods~\cite{sohn2020fixmatch}, we utilize the confident predictions of each detector on easy views, to enforce consistency of the other detector on hard views. The overall loss function consists of the heatmap regression loss $\mathcal{L}_{heat}$ on the exemplar and selected samples, and the cross-consistency loss $\mathcal{L}_{con}^{cross}$ between the easy-hard pairs. 

\section{Experiments}
We evaluate our method and state-of-the-art methods on multiple public X-ray datasets of different body parts. Furthermore, we conduct ablation studies to demonstrate how different components contribute to our final performance. 

\subsection{Dataset}
\subsubsection{Head:} This dataset is a widely-used open-source dataset collected for IEEE ISBI 2015 challenge~\cite{wang2015evaluation,wang2016benchmark}, which consists of 400 cephalometric radiographs. Two medical experts annotate 19 landmarks manually and we compute their average as the ground truth like~\cite{payer2019integrating,yao2021one}. The entire dataset is officially split into three parts: the first 150 images are training set, the next 250 images are test set. The original resolution is $2400\times 1935$ and the pixel spacing is 0.1 mm.

\subsubsection{Hand:} This dataset is a public dataset of hand radiographs collected by \cite{gertych2007bone}. \cite{payer2019integrating} further annotate 37 landmarks on fingertips and bone joints. They assume the length between two endpoints of the wrist is 50 mm. The whole dataset is split into a training set of 609 images and a test set of 300 images as in \cite{zhu2021you}. 

\subsubsection{Chest:} This dataset is from a Kaggle challenge. \cite{zhu2021you} select a subset of 279 images by excluding abnormal cases and annotate six landmarks at the boundaries of the lung. Following \cite{zhu2021you}, we use pixel distance at fixed resolution ($512\times512$) for evaluation, since no pixel spacing information is provided. 

\subsection{Implementation}
For both stages, the input resolution for head and hand datasets is $320\times 256$, while chest images are resized into $256\times 256$ to keep the aspect ratio. Data augmentations used in both stages include random rotation and random scaling. Random horizontal flipping is only applied in stage \uppercase\expandafter{\romannumeral2}.
$\beta_1, \beta_2$ for Adam optimizer is set to $0.99, 0.0$ respectively and weight decay is 1e-4.

For stage \uppercase\expandafter{\romannumeral1}, we adopt pretrained HRNet18~\cite{wang2020deep} as backbone. The channel dimension of extracted feature maps are adjusted to $d_{model}=64$ through $1\times 1$ convolution, as the input dimension for attention blocks. Self (cross) attention modules are implemented as the encoder (decoder) block designed in~\cite{vaswani2017attention}, with multi-head attention ($N_h=2$) and dropout $p=0.1$. For each step of deformation, we stack $N_l=2$ attention blocks for feature fusion. For alignment estimation, we set $sf_x, sf_y, rot, sh$ to $1, 1, \frac{\pi}{2}, \frac{\pi}{2}$ respectively. We set $\sigma=3$ for heatmap generation in Eq.~\ref{esim} and $T=0.1$ in Eq.~\ref{esmooth}.

We train the model for 750 epochs. For the first 250 epochs, learning rate $lr$ is fixed as 1e-4 and ramps the weight for local deformation $\lambda_1$ (Eq.~\ref{loss_st1}) from $0$ to $1$. For the remaining epochs, $lr$ is decayed to 5e-5 with the cosine annealing strategy. In Eq.~\ref{loss_reg}, $\lambda_2$ is set as $0.25$ and $\lambda_3$ cosinely ramps down from $5.0$ to $0.0$. For each mini-batch, half of them are synthesized source-target pairs with known pixel correspondence, while the others are obtained by shuffling images within the current batch. In consideration of computation overhead and memory, all models are trained with one global step and two subsequent local steps ($N_\Phi=2$). We start to infer pseudo labels for the training set after 200 epochs and compute its exponential moving average with a $\tau$ of 0.9 in Eq.~\ref{ema}.

For stage \uppercase\expandafter{\romannumeral2}, we retrain two HRNet18~\cite{wang2020deep} detectors, both initialized from pre-trained weights. We set $\sigma=3$ for the standard deviation of the gaussian heatmap. Initial $lr$ is 1e-3 and decays by $0.1$ at 60 epochs and 80 epochs, until a total of 100 epochs. In the first 30 epochs, filter rate $\epsilon$ ramps from 0.0 to 0.8.

\subsection{Evaluation}
\begin{table}[htbp]\scriptsize\centering
\renewcommand{\tabcolsep}{1.0pt}\renewcommand{\arraystretch}{1.2}
\caption{Evaluation on the head, hand and chest dataset. * denotes original reported results in paper. $\#$ denotes reproduced performances under the one-shot landmark setting. Supervised method YOLO~\cite{zhu2021you} serves as an universal upper bound. The best results are in \textbf{bold} and the second best results are \underline{underlined}. Strictly using the same exemplar, our result outperforms all other one-shot methods. Performance of stage \uppercase\expandafter{\romannumeral1} is reported for fair comparison with CC2D-SSL}
\resizebox{.975\linewidth}{!}{
\begin{tabular}{l | ccccc | cccc | cccc}
\toprule[1pt]
\multirow{3}{*}{\bf{Method}} & \multicolumn{5}{c|}{\bf{Head}} & \multicolumn{4}{c|}{\bf{Hand}} & \multicolumn{4}{c}{\bf{Chest}} \\
& MRE$\downarrow$ & \multicolumn{4}{c|}{SDR$\uparrow$(\%)} & MRE$\downarrow$ & \multicolumn{3}{c|}{SDR$\uparrow$(\%)} & MRE$\downarrow$ & \multicolumn{3}{c}{SDR$\uparrow$(\%)}\\
        & (mm) & 2mm   & 2.5mm  & 3mm    & 4mm    & (mm) & 2mm  & 4mm   & 10mm & (px) & 3px & 6px & 9px \\
\hline
YOLO$*$    & 1.54 & 77.79 & 84.65  & 89.41  & 94.93  & 0.84 & 95.4 & 99.35 & 99.75 & 5.57 & 57.33 & 82.67 & 89.33 \\
\hline\hline
3FabRec$\#$ & 20.12 & 2.42 & 3.86 & 4.98 & 7.23          & 9.81          & 3.98           & 15.24          & 60.92          & 48.67         & 0.67           & 2.33           & 4.67           \\
\hline
DataAug$\#$ & 3.18 & 32.81 & 44.42 & 55.12 & 73.16 & 2.51 & 48.87 & 85.67 & 98.91 & \ul{10.15} & \ul{15.67} & \ul{40.67} & \ul{61.67}  \\
\hline
CC2D-SSL$\#$  & 3.41 & 40.63 & 49.58 & 60.31 & 72.14 & 2.93 & 51.59 & 81.29 & 95.59 & 17.37 & 9.87 & 27.99 & 42.11 \\
CC2D-TPL$\#$  & \ul{2.72} & 42.59 & 53.18 & \ul{66.48} & \ul{83.22} & 2.47 & 54.95 & 87.16 & 97.84 & 12.91 & 12.67 & 38.67 & 57.67 \\
\hline
Ours-stage \uppercase\expandafter{\romannumeral1} & 2.70 & \ul{42.78} & \ul{54.88} & 65.03 & 81.01 & \ul{2.13}  & \ul{60.93} & \ul{89.43}  & \ul{99.21} & 10.16 & 12.33 & 39.00 & 60.33 \\
Ours-stage \uppercase\expandafter{\romannumeral2} & \bf{2.13} & \bf{54.69} & \bf{67.47} & \bf{77.85} & \bf{90.02} & \bf{1.82} & \bf{66.39} & \bf{92.93} & \bf{99.97} & \bf{6.89} & \bf{17.33} & \bf{50.33} & \bf{75.33} \\
\bottomrule[1pt]
\end{tabular}
}
\label{total}
\end{table}

\subsubsection{Metrics}
Mean radial error (MRE) and successful detection rate (SDR) are adopted as evaluation metrics. MRE computes the average Euclidean distances between predicted landmarks and ground truth landmarks. Given several thresholds, SDR calculates the proportion of predictions with an error below these thresholds respectively. The unit of MRE is mm if pixel spacing is provided. Otherwise, it is reported as raw pixel distance. We use the same thresholds for SDR as ~\cite{zhu2021you}, where they developed a fully-supervised universal landmark detector trained on the mix of all three datasets aforementioned. We list their results in Tab.~\ref{total} as the supervised upper bound.

\subsection{Comparison with Baseline Methods}
As shown in Tab.~\ref{total}, we compare with 3FabRec, DataAug and CC2D on all three datasets. Our results outperform other one-shot methods by notable margins, which demonstrates both effectiveness and general applicability of our method. 

\begin{figure*}[t]\centering
\includegraphics[width=.92\textwidth]{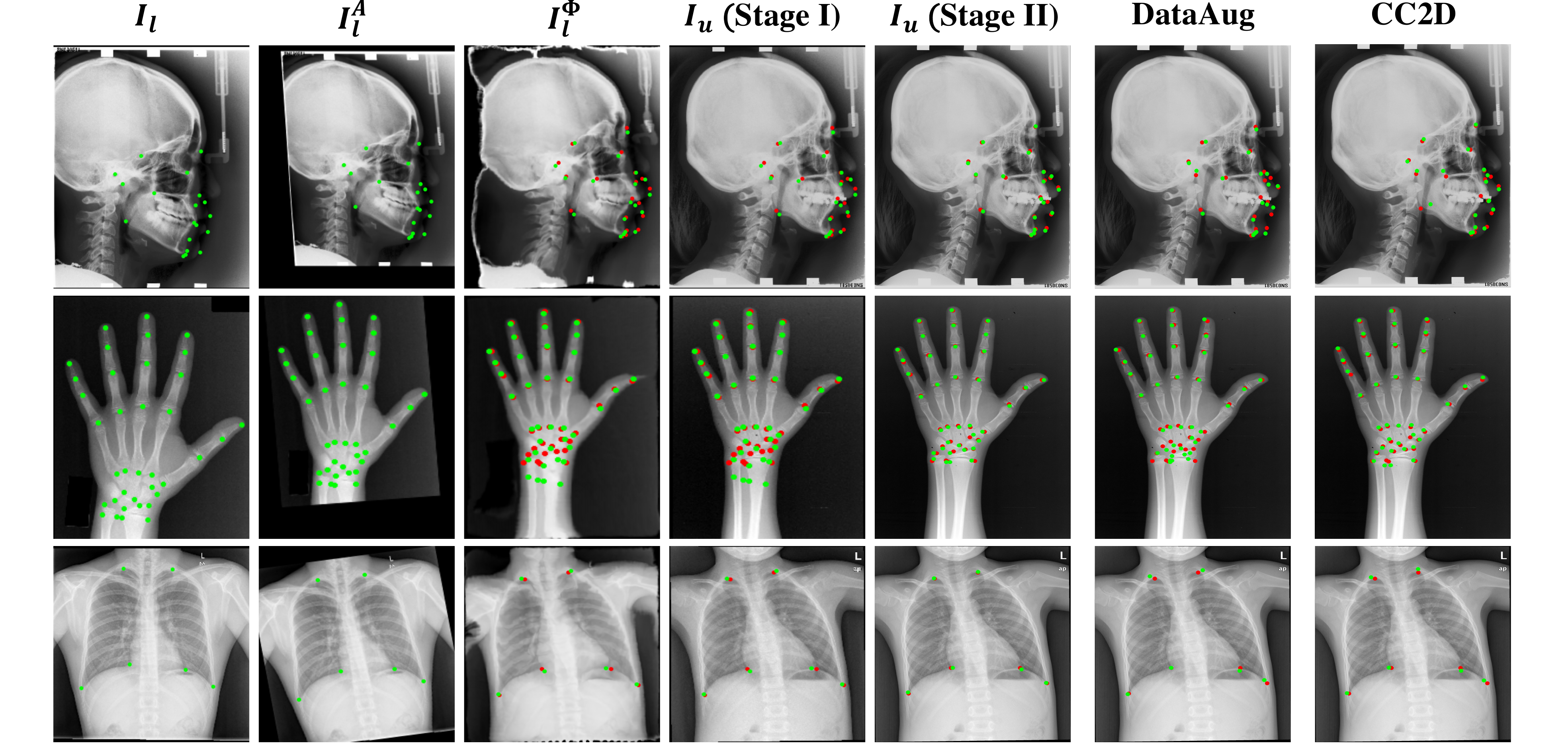} 
\caption{Visualization of learned transformations and comparison with baseline methods. From left to right, we display the exemplar ($I_{src}$), intermediate warped results ($I_{l}^A, I_{src}^{\Phi}$) and the unlabeled target image $I_{u}$. \textbf{Green dots} denote locations of the transformed exemplar landmarks. \textbf{Red dots} denote the ground truth landmark locations of $I_{u}$. Most of the stage \uppercase\expandafter{\romannumeral1} predictions with large biases could be corrected in stage \uppercase\expandafter{\romannumeral2}}
\label{showcase}
\end{figure*}

As the weak baseline, with only access to hundreds of images, 3FabRec can hardly reconstruct the fine-grained details in X-ray images and thus yield relatively poor results. Our method also shows consistent improvements against the strong baseline DataAug, which relies on the affine alignment in the pre-processing step. While for some cases with drastic changes in spatial structure as in Fig.~\ref{showcase}, our end-to-end learned alignment is more beneficial for the subsequent local deformations, since the total transformation learning is extraly guided by our carefully-designed constraints for landmarks. 

Compared to the state-of-the-art method CC2D, we achieve superior performance by decreasing the MRE of stage \uppercase\expandafter{\romannumeral1} and stage \uppercase\expandafter{\romannumeral2} by $20.8\%$ (3.41mm$\rightarrow$2.70\ mm) and $21.7\%$ (2.72mm$\rightarrow$2.13mm) respectively. Improvements over CC2D might be attributed to the following two aspects. First, the pretext task of image-patch matching used in~\cite{yao2021one}, is prone to overfitting when having difficulties discriminating local regions centered on those densely-labeled landmarks. Besides, they do not consider the global spatial relationships among these patches. In contrast, our way of inferring landmarks by registration, implicitly takes such constraints into consideration, and thus naturally avoid abnormal predictions, which can be justified by the obvious increase of SDR@4mm in stage \uppercase\expandafter{\romannumeral1} ($72.14\% \rightarrow 81.01\%$). Second, compared to simply retraining with synthesized samples as in DataAug, or performing majority-voting as in CC2D, our C2T makes better use of the common landmark knowledge contained in pseudo labels. Specifically, reliable labels can be effectively selected to provide more supervision for heatmap regression. On the other hand, remaining instances are not wasted by contributing to the consistency learning. It is especially crucial to handle challenging cases, such as the second row in Fig.~\ref{showcase}, which successfully correct the large biases introduced in stage \uppercase\expandafter{\romannumeral1}, for landmarks around wrists. 

\begin{table*}[htbp]\scriptsize \centering
\renewcommand{\tabcolsep}{1.0pt}\renewcommand{\arraystretch}{1.5}
\parbox{.455\textwidth}{
\caption{Ablation of spatial transform. ``L" denotes the alignment is learned}
    \resizebox{1.0\linewidth}{!}{
    \begin{tabular}{l | c | ccccc}
    \toprule[1pt]
    \bf{Global}                  & \bf{Local}        & \multicolumn{5}{c}{\bf{Head}}  \\
    \bf{Alignment}                   & \bf{Step}         & MRE$\downarrow$  & \multicolumn{4}{c}{SDR$\uparrow$(\%)}       \\
    \bf{Type}                    & $N_\Phi$          & (mm) & 2mm   & 2.5mm & 3mm   & 4mm   \\
    \hline
    \xmark                &  2   & 3.42 & 30.78 & 41.96 & 52.02 & 69.36 \\
    sift                  &  2   & 3.10 & 35.31 & 47.58 & 58.23 & 74.02 \\
    affine (L)      &  1   & 2.73 & 39.64 & 52.19 & 63.45  & 80.50 \\
    affine (L)       &  2   & 2.70 & 42.78 & 54.88 & 65.03 & 81.01 \\
    \hline 
    perspective (L)   &  2   & 2.80 & 40.17 & 52.48 & 63.41 & 79.35 \\
    \bottomrule[1pt]
    \end{tabular}
    \label{abl_transform}
    }
}
\quad
\parbox{.475\textwidth}{
\caption{Configurations of stage \uppercase\expandafter{\romannumeral1} network}
    \resizebox{1.0\linewidth}{!}{
    \begin{tabular}{l | c | c | ccccc}
    \toprule[1pt]
    \multirow{3}{*}{\bf{backbone}} & \multirow{3}{*}{\bf{$N_l$}} & \multirow{3}{*}{\bf{$N_h$}} & \multicolumn{5}{c}{\bf{Head}}    \\
                              &                     &                     & MRE$\downarrow$  & \multicolumn{4}{c}{SDR$\uparrow$(\%)}\\
                              &                     &                     & (mm) & 2mm   & 2.5mm & 3mm   & 4mm   \\
    \hline 
    \multirow{4}{*}{hrnet}    & 0                   & 0                   & 2.93 & 36.76 & 48.36 & 59.58 & 76.11 \\
                              & 1                   & 1                   & 2.85 & 37.37 & 50.08 & 60.67 & 78.36 \\
                              & 2                   & 1                   & 2.82 & 39.87 & 52.61 & 63.47 & 78.25 \\
                              & 2                   & 2                   & 2.70 & 42.78 & 54.88 & 65.03 & 81.01 \\
    \hline 
    unet                      & 2                   & 2                   & 2.95 & 34.86 & 47.81 & 58.63 & 76.42 \\
    \bottomrule[1pt]
    \end{tabular}
    \label{abl_model}
    }
}
\end{table*}

\subsection{Ablation Study}
 Here we conduct experiments on the head dataset, to analyze the contributions of different components from multiple aspects including spatial transformation, network architecture, loss function and robustness to exemplar selection.
 
\subsubsection{Spatial Transformation}
We remove the global alignment in the first row of Tab.~\ref{abl_transform} and witness a great drop in our performance. Besides, compared to alignment computed by traditional methods (e.g., sift detector), learned alignment is more desirable to reduce the complexity for subsequent local deformations. Compared to single step of local deformation, one more step benefits our performance, decreasing MRE from 2.73mm to 2.70mm. This observation is well-aligned with ~\cite{zhao2019recursive}. We also try to learn perspective transform with more degrees of freedom, but do not observe further improvements as expected. We argue that this might be restricted by the complexity of the head dataset. 

\subsubsection{Stage \uppercase\expandafter{\romannumeral1} Network Configurations}
We remove all attention blocks in the first row of Tab.~\ref{abl_model} and achieve a MRE of 2.93mm. Furnishing backbone with more layers of attention blocks ($N_l=0\rightarrow2$) and multi-head attention ($N_h=1\rightarrow2$) contributes positively to our final performance. Our method still achieves competitive performance if we switch our backbone to the commonly-used UNet~\cite{ronneberger2015u}. 

\begin{table*}[htbp]\scriptsize\centering
\renewcommand{\tabcolsep}{1.0pt}\renewcommand{\arraystretch}{1.2}
\parbox{.455\textwidth}{
\caption{Ablation of loss function for stage \uppercase\expandafter{\romannumeral1}}
\resizebox{1.0\linewidth}{!}{
\begin{tabular}{l | ccccc}
\toprule[1pt]
\multirow{3}{*}{\bf{Loss}} & \multicolumn{5}{c}{\bf{Head}}    \\
                               & MRE$\downarrow$ & \multicolumn{4}{c}{SDR$\uparrow$(\%)} \\
                               & (mm)   & 2mm   & 2.5mm & 3mm   & 4mm   \\
\hline
w/o $\mathcal{L}_{inv}$     & 3.24 & 32.67 & 44.95 & 55.52 & 71.81  \\
w/o $\mathcal{L}_{smooth}$    & 2.97 & 39.01 & 50.23 & 60.46 & 75.85 \\
w/o $\mathcal{L}_{syn}$     & 2.86 & 39.54 & 51.43 & 61.68 & 77.14 \\
\hline 
$\mathcal{L}_{esim} \rightarrow \mathcal{L}_{sim} $  & 3.17 & 33.31 & 45.64 & 57.31 & 73.01 \\
$\mathcal{L}_{esmooth} \rightarrow \mathcal{L}_{smooth}$  & 2.75 & 40.27 & 52.95 & 64.65 & 81.12 \\
ours                       & 2.70 & 42.78 & 54.88 & 65.03 & 81.01 \\
\bottomrule[1pt]
\end{tabular}
\label{loss1}
}
}
\qquad
\parbox{.475\textwidth}{
\caption{Ablation for stage \uppercase\expandafter{\romannumeral2}} 
\resizebox{1.0\linewidth}{!}{
\begin{tabular}{l | c |ccccc}
\toprule[1pt]
\multirow{3}{*}{\bf{$\mathcal{L}_{filter}$}} & \multirow{3}{*}{\bf{$\mathcal{L}^{cross}_{con}$}} & \multicolumn{5}{c}{\bf{Head}}             \\
        &             & MRE$\downarrow$ & \multicolumn{4}{c}{SDR$\uparrow$(\%)}       \\
        &             & (mm) & 2mm   & 2.5mm & 3mm   & 4mm   \\
\hline
\xmark                           & \xmark   & 2.53 & 44.38 & 57.37 & 67.52 & 83.43 \\
w $\mathcal{L}^{self}_{con}$  & \xmark   & 2.29 & 50.80 & 63.64 & 74.36 & 87.53 \\
w/o $\mathcal{L}^{self}_{con}$  & \cmark   & 2.17 & 53.81 & 67.18 & 77.43 & 89.70 \\
w $\mathcal{L}^{self}_{con}$     & \cmark   & 2.13 & 54.69 & 67.47 & 77.85 & 90.02 \\
\bottomrule[1pt]
\end{tabular}
\label{loss2}
}
}
\end{table*}

\subsubsection{Loss Function for Stage\uppercase\expandafter{\romannumeral1}}
In the first three rows of Tab.~\ref{loss1}, we remove one regularization term of $\Phi$ at a time and find they all contribute to our results. They require the deformation field to be smooth, reversible, and applicable to real images respectively, thus avoiding intermediate abnormal warping results.

To study the contributions of $\mathcal{L}_{esim}$, we replace it with $\mathcal{L}_{sim}$ and find that MRE increases greatly by $17.4\%$. $\mathcal{L}_{sim}$ only considers image similarity based on pixel values, while our $\mathcal{L}_{esim}$ further incorporates the local structural similarity around landmarks based on edge maps, which is crucial to enforce the consistency across instances and thus adapts better to landmark localization task. Compared to $\mathcal{L}_{esmooth}$, $\mathcal{L}_{smooth}$ also leads to slight degradation in performance. Taking landmarks annotated near boundaries between foreground and background for example, they might be biased towards background due to the strict smoothness enforced by $\mathcal{L}_{smooth}$. $\mathcal{L}_{esmooth}$ is tolerant to discontinuity between different regions and thus mitigates such phenomenon. 

\subsubsection{Ablation for Stage \uppercase\expandafter{\romannumeral2}}
As in Tab.~\ref{loss2}, we achieve a MRE of 2.53mm by simply retraining with pseudo labels in stage \uppercase\expandafter{\romannumeral1}. Filtering out noisy pseudo labels by heatmap loss $\mathcal{L}_{heat}$ can improve MRE to 2.29mm. And $\mathcal{L}_{con}^{cross}$ involves these filtered samples for semi-supervised learning, which further decreases MRE to 2.17mm. If $\mathcal{L}_{con}^{self}$ is also taken into consideration for filtering, selected pseudo labels can provide more reliable supervision since they are also robust against spatial transformations, which enables us to achieve the final MRE of 2.13mm. 

\begin{figure}[t]\centering
\includegraphics[width=.95\textwidth]{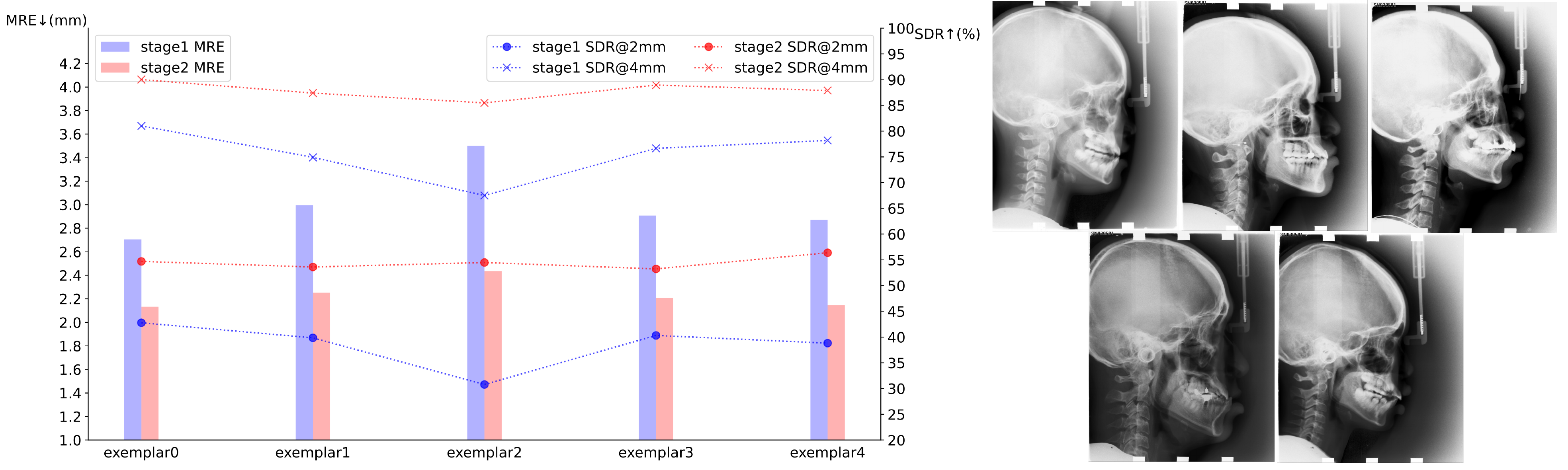}
\caption{Experiments for five candidate exemplars on the head dataset}
\label{exemplar}
\end{figure}
\subsubsection{Exemplar Selection}
In Fig.~\ref{exemplar}, we show five different candidate exemplars and their corresponding results on the head dataset. Although there might be certain variations in the performances of stage \uppercase\expandafter{\romannumeral1}, our retraining scheme introduced in stage \uppercase\expandafter{\romannumeral2} effectively stabilizes the final performance, with a mean MRE of 2.25mm.

\section{Conclusion}
We propose a novel two-stage framework for one-shot medical landmark localization. In stage \uppercase\expandafter{\romannumeral1}, an image transform model is learned through unsupervised registration. The total transform is decomposed to an end-to-end cascade of global alignment and local deformations, with the guidance of novel loss functions incorporating edge information. Pseudo landmarks are inferred on unlabeled targets with the exemplar and learned transform model. In stage \uppercase\expandafter{\romannumeral2}, we use these noisy labels to train robust landmark detectors by exploring self-consistency for selecting reliable samples and cross-consistency for semi-supervised learning. Extensive experiments on multiple datasets demonstrate our method surpasses other one-shot methods and further narrows the gap with supervised methods.

\noindent\textbf{Acknowledgements} This work is supported in part by MOST-2018AAA010200\\4 and NSFC-62061136001.

\clearpage
%
%
\bibliographystyle{unsrt}
\bibliography{egbib}
\end{document}